\documentclass{article}
\usepackage{ijcai26}

\usepackage{times}
\usepackage{soul}
\usepackage{url}
\usepackage[hidelinks]{hyperref}
\usepackage[utf8]{inputenc}
\usepackage[small]{caption}
\usepackage{graphicx}
\usepackage{amsmath}
\usepackage{amsthm}
\usepackage{amssymb}
\usepackage{booktabs}
\usepackage{algorithm}
\usepackage{algorithmic}
\usepackage[switch]{lineno}

\usepackage{graphicx}

\usepackage{makecell}
\usepackage{colortbl}
\usepackage{mathrsfs}
\usepackage{multirow}
\usepackage{booktabs}

\title{GPD: Guided Progressive Distillation for Fast and High-Quality Video Generation}

\author{
    Xiao Liang
    \and
    Yunzhu Zhang
    \and
    Linchao Zhu\thanks{Corresponding Author.} \\ 
    \affiliations
    College of Computer Science and Technology, Zhejiang University \\
    \emails
    \{0314liangxiao, yunzhuzhang0918, zhulinchao7\}@gmail.com
}

\begin{document}
\maketitle

\begin{abstract}
Diffusion models have achieved remarkable success in video generation; however, the high computational cost of the denoising process remains a major bottleneck. Existing approaches have shown promise in reducing the number of diffusion steps, but they often suffer from significant quality degradation when applied to video generation. We propose Guided Progressive Distillation (GPD), a framework that accelerates the diffusion process for fast and high-quality video generation.
GPD introduces a novel training strategy in which a teacher model progressively guides a student model to operate with larger step sizes. The framework consists of two key components: (1) an online-generated training target that reduces optimization difficulty while improving computational efficiency, and (2) frequency-domain constraints in the latent space that promote the preservation of fine-grained details and temporal dynamics.
Applied to the Wan2.1 model, GPD reduces the number of sampling steps from 48 to 6 while maintaining competitive visual quality on VBench. Compared with existing distillation methods, GPD demonstrates clear advantages in both pipeline simplicity and quality preservation.
\end{abstract}

\section{Introduction}

Diffusion models \cite{hoDenoisingDiffusionProbabilistic2020b,songScoreBasedGenerativeModeling2021} have emerged as a powerful framework for video generation \cite{blattmannStableVideoDiffusion2023,zhouMagicVideoEfficientVideo2023,wanWanOpenAdvanced2025a,wangMagicVideoV2MultiStageHighAesthetic2024,kongHunyuanVideoSystematicFramework2025,maStepVideoT2VTechnicalReport2025}, achieving state-of-the-art results in temporal consistency and visual quality.
However, video diffusion models require iterative denoising of Gaussian noise to generate final videos, making them computationally intensive and time-consuming.

\begin{figure}[t]
    \centering
    \includegraphics[width=\linewidth]{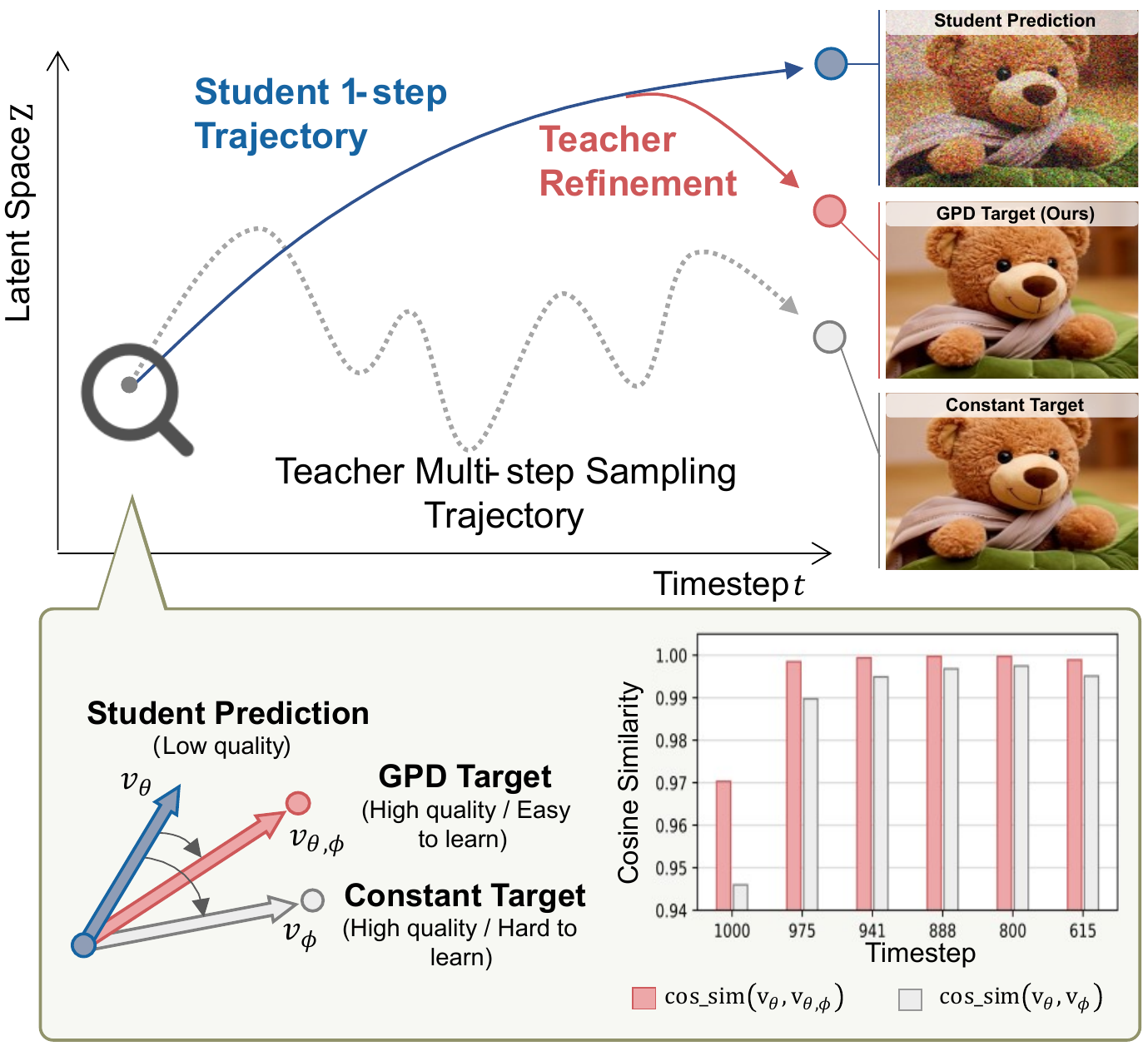}
    \caption{Motivation of Guided Progressive Distillation (GPD). Traditional flow-straightening forces a student to approximate the teacher’s highly curved multi-step sampling trajectory in a single step, often causing large target mismatch and low-quality predictions. We propose teacher refinement, where the teacher further refines the student’s one-step intermediate output to form an adaptive GPD target that is both high-fidelity and easier to learn than a constant target. As shown by the vector alignment analysis and cosine similarity across timesteps, the refined target’s update direction aligns better with the student’s current inference flow.}
    \label{fig:intro}
\end{figure}

To accelerate the diffusion process, traditional flow-straightening methods~\cite{liuFlowStraightFast2022a,liuInstaFlowOneStep2024,yanPeRFlowPiecewiseRectified2024b,keProReflowProgressiveReflow2025} force the student to cover the highly curved multi-step trajectory of the teacher in one step. This objective is inherently challenging and often leads to poor alignment, where the student's one-step prediction deviates significantly from the teacher's target, resulting in low-quality outputs as shown in Figure~\ref{fig:intro}. 
We find that incorporating an additional teacher refinement step, based on the student's one-step intermediate prediction, can be beneficial. Specifically, it offers two key advantages. First, the refinement step produces high-quality supervision targets. By refining the student's initial prediction, the teacher provides high-fidelity signals that guide the student to produce videos with superior visual details and temporal consistency. Second, compared to the constant targets used in conventional methods, the teacher-refined target is adaptive and therefore easier to learn. As evidenced by the higher cosine similarity in Figure \ref{fig:intro}, the update direction of the refined target aligns more consistently with the student's current inference flow. In summary, this teacher refinement strategy effectively bridges the gap between generation quality and training difficulty, ensuring that the student learns from a target that is both high-fidelity and structurally aligned with its inference path.

Based on this insight, we further extend the concept of progressive optimization~\cite{linSDXLLightningProgressiveAdversarial2024,salimansProgressiveDistillationFast2022a} and incorporate quality-preserving techniques, proposing Guided Progressive Distillation (GPD), an acceleration framework that achieves efficient, high-quality video generation.
First, we introduce online refinement training, where the teacher model dynamically refines the student's intermediate predictions during training, creating adaptive targets that follow straighter trajectories compared to teacher synthesis.
Second, we design a new progressive distillation strategy adapted to online refinement training, enabling fast and stable convergence. Compared to other distillation methods, it substantially reduces training cost.
Third, we incorporate high-frequency preservation through a frequency-domain loss that explicitly maintains fine-grained details and motion dynamics that are often smoothed out during trajectory straightening.

Extensive experiments demonstrate that GPD achieves state-of-the-art performance, generating high-quality videos in just 6 steps compared to the original 48 steps---an 8$\times$ speedup. On the VBench benchmark~\cite{huangVBenchComprehensiveBenchmark2023}, our method achieves a total score of 84.04\%, outperforming both traditional full-step models and existing acceleration methods including CausVid~\cite{yinSlowBidirectionalFast2025} (83.65\%) and AccVideo~\cite{zhangAccVideoAcceleratingVideo2025a} (83.28\%), while maintaining superior text alignment and visual fidelity across diverse video generation tasks.


\section{Related Work}

Video diffusion distillation methods can be broadly classified into two categories according to training objective: Distribution Matching methods and Trajectory Straightening methods.

Distribution Matching methods aim to align the output distribution of the student model with that of the teacher model.
These methods leverage the score function to minimize the divergence of the distributions between teacher and student output~\cite{luoOneStepDiffusionDistillation2024,yinOnestepDiffusionDistribution2024a,chadebecFlashDiffusionAccelerating2024,sauerAdversarialDiffusionDistillation2023,yinImprovedDistributionMatching2024,linSDXLLightningProgressiveAdversarial2024}.
Recent advances include~\cite{yinSlowBidirectionalFast2025}, which utilizes a teacher-synthesized dataset to train a one-step generator and extends Distribution Matching Distillation (DMD) as a post-training stage to enhance generation quality.
However, these approaches rely on real video datasets for distribution matching, and the generated captions often lack fine-grained detail, which can weaken text–video alignment and reduce controllability.

Trajectory Straightening methods aim to straighten the probability flow trajectory by forcing the student model to predict the same target as the teacher's multi-step prediction.
Consistency Model~\cite{songConsistencyModels2023a} leverages a novel training paradigm that enables the model to predict the initial image at any timestep of the diffusion process, ultimately enabling training of a one-step generator. ~\cite{wangVideoLCMVideoLatent2023,luoLCMLoRAUniversalStableDiffusion2023} extended Consistency Model to latent space, facilitating high-resolution image and video synthesis with minimal inference steps.

Flow-based methods~\cite{lipmanFlowMatchingGenerative2023a} accelerate diffusion models by straightening the velocity field predicted by diffusion model.~\cite{liuFlowStraightFast2022a,liuInstaFlowOneStep2024} introduce Rectified Flow to enhance both efficiency and transferability in data generation tasks. The concept was further expanded by~\cite{esserScalingRectifiedFlow2024a,yanPeRFlowPiecewiseRectified2024b,keProReflowProgressiveReflow2025}, enabling efficient high-resolution image synthesis through transformer architectures.
While these methods are easy to train using a simple regression loss, they often result in blurry outputs during few-step generation, especially when applied to video synthesis.

\section{Background}
\subsection{Flow Matching for Diffusion Models}

Diffusion models generate data by learning to reverse a gradual noising process~\cite{hoDenoisingDiffusionProbabilistic2020b,songScoreBasedGenerativeModeling2021}. Under the flow-matching paradigm~\cite{lipmanFlowMatchingGenerative2023a}, this process is formulated as learning a time-dependent velocity field $v_\theta(z_t, t)$ that transports samples along a probability flow ODE:
\begin{equation}
\frac{dz_t}{dt} = v_\theta(z_t, t), \quad t \in [0, T],
\end{equation}
where $z_t$ represents the noisy data at time $t$, with $z_0$ being clean data and $z_T$ being pure noise.

In practice, solving the ODE requires discretization. Let $\mathcal{S}_\theta(z_{t_i}, t_i, t_j)$ denote the ODE solver that integrates the velocity field $v_\theta$ from time $t_i$ to $t_j$:
\begin{equation}
\begin{aligned}
\mathcal{S}_\theta(z_{t_i}, t_i, t_j)
&= z_{t_i} + \int_{t_i}^{t_j} v_\theta(z_t, t)\, dt \\
&\approx z_{t_i} + (t_j - t_i)\cdot v_\theta(z_{t_i}, t_i).
\end{aligned}
\end{equation}

For a teacher model with parameters $\phi$, the generation process consists of $N$ sequential denoising steps:
\begin{equation}
\begin{aligned}
z_0
&= \mathcal{S}_\phi^{(N)}(z_{t_N}) \\
&= \mathcal{S}_\phi(z_{t_N}, t_N, t_{N-1}) \circ \cdots \circ \mathcal{S}_\phi(z_{t_1}, t_1, t_0).
\end{aligned}
\end{equation}

High-quality generation typically requires many evaluations (e.g., 50-100 steps) of the velocity field, which becomes computationally prohibitive, especially for video generation.

\subsection{Distillation for Acceleration}

Distillation methods aim to train a student model with parameters $\theta$ that can generate high-quality samples in $K \ll N$ steps. The rectified flow distillation approach, which we refer to as offline distillation, attempts to directly match the student's $k$-step prediction with pre-computed teacher trajectories~\cite{zhangAccVideoAcceleratingVideo2025a}:

\begin{equation}
\mathcal{L}_{\text{standard}} = \mathbb{E}_{z_{t_i}} \left\| \mathcal{S}_\theta(z_{t_i}, t_i, t_{i-k}) - \mathcal{S}_\phi^{(k)}(z_{t_i}) \right\|^2,
\end{equation}

where $\mathcal{S}_\phi^{(k)}(z_{t_i})$ denotes the teacher's $k$-step trajectory.

\begin{figure*}[t]
    \centering
    \includegraphics[width=\textwidth]{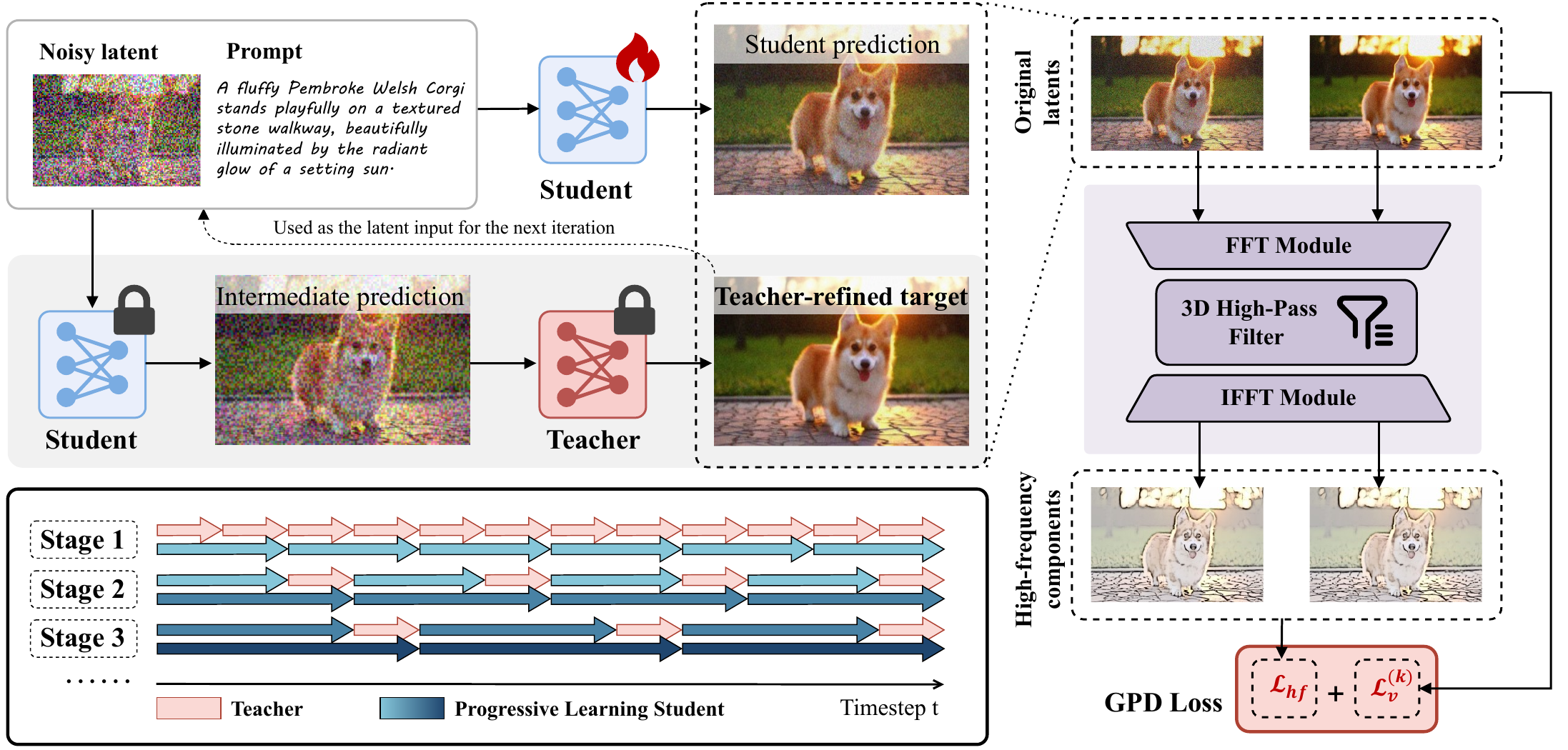}
    \caption{Overview of Guided Progressive Distillation. \textbf{Progressive distillation:} training is split into stages that gradually extend the student’s step size. \textbf{Teacher-refined target:} the teacher refines an intermediate latent predicted by the frozen prior student to form the supervision target. \textbf{High-frequency loss:} a 3D FFT high-pass penalty enforces matching high-frequency details between the student output and the teacher-refined target.}
    \label{fig:main}
\end{figure*}

\section{Method}
\subsection{Guided Progressive Distillation}

Traditional distillation approaches employ the loss $\mathcal{L}_{\text{standard}}$ with pre-computed teacher trajectories. The student model is required to match fixed targets sampled by the teacher along a highly curved trajectory in the latent space. This formulation suffers from two major shortcomings: (1) when $k$ is large, the teacher trajectory becomes highly complex due to the accumulation of nonlinear transformations. Such high-curvature trajectories are difficult for the student to approximate within a single prediction step; (2) during training, inputs are generated solely by the teacher. At inference, errors from intermediate student predictions may accumulate, causing a distribution shift that remains unaddressed.

\textbf{Progressive distillation with linear increment.} To address the challenge of high trajectory curvature, we decompose the optimization process into $K$ progressive stages, linearly increasing the prediction horizon of the student model. Assuming both the teacher and student share an initial unit step size, at each stage $k \in {2, ..., K}$, we train the student to approximate a target with a step size of $k$. By gradually increasing the step size from 1 to $K$ across multiple stages, this approach ensures a smooth transition in trajectory curvature, making the learning landscape tractable. 

Instead of requiring the student to immediately bridge a complex $K$-step gap involving highly nonlinear trajectories, the progressive strategy enables the model to incrementally learn longer predictions. Each stage builds upon the stable foundation established by the previous one, allowing the student to adapt to increasing complexity in a structured and efficient manner.

\textbf{Online training target with teacher guidance.} To address the distribution shift between training and inference, we propose a dynamic online target generation mechanism. Instead of relying on static, precomputed teacher trajectories, our method constructs training targets dynamically via a three-model interaction process. This design allows the training objective to adapt continuously to the evolving capabilities of the student model. 

Specifically, we use a frozen copy of the student model, $v_{\theta_{k-1}}$, from the previous training stage to estimate the velocity field over a $k-1$-step interval from the initial latent state $z_{t_i}$. An ODE solver integrates this velocity to yield an intermediate latent $z_{t_{i-k+1}}$. 
\begin{equation}
z_{t_{i-k+1}} = \mathcal{S}_{\theta_{k-1}}(z_{t_i}, t_i, t_{i-k+1}).
\end{equation}

This intermediate state is then passed to the teacher model $v_\phi$, which refines the trajectory by computing the velocity for the final step.
\begin{equation}
z_{t_{i-k}}^* = \mathcal{S}_\phi(z_{t_{i-k+1}}, t_{i-k+1}, t_{i-k}).
\end{equation}

The resulting refined latent state $z_{t_{i-k}}^*$ is used to compute a global average velocity over the full $k$-step interval:
\begin{equation}
v_{\text{target}} = \frac{z_{t_{i-k}}^* - z_{t_i}}{t_{i-k} - t_i}.
\end{equation}
This dynamically derived velocity serves as the optimization target for the current student model $v_\theta$. Such online target generation enhances robustness by enabling the teacher to correct the student's trajectory progressively.

The training objective for trajectory straightening at stage k is defined as:
\begin{equation}
\mathcal{L}_v^{(k)} = \mathbb{E}_{z_{t_i}, t_i} \left\| v_\theta(z_{t_i}, t_i) - \frac{z_{t_{i-k}}^* - z_{t_i}}{t_{i-k} - t_i} \right\|^2.
\end{equation}
This progressive, teacher-refined training pipeline combines the benefits of gradual time-step extension with online, adaptive supervision. As a result, the student learns to generalize across a wider temporal horizon, with improved resilience to distributional discrepancies between training and inference.

\begin{figure}[t]
    \centering
    \includegraphics[width=\linewidth]{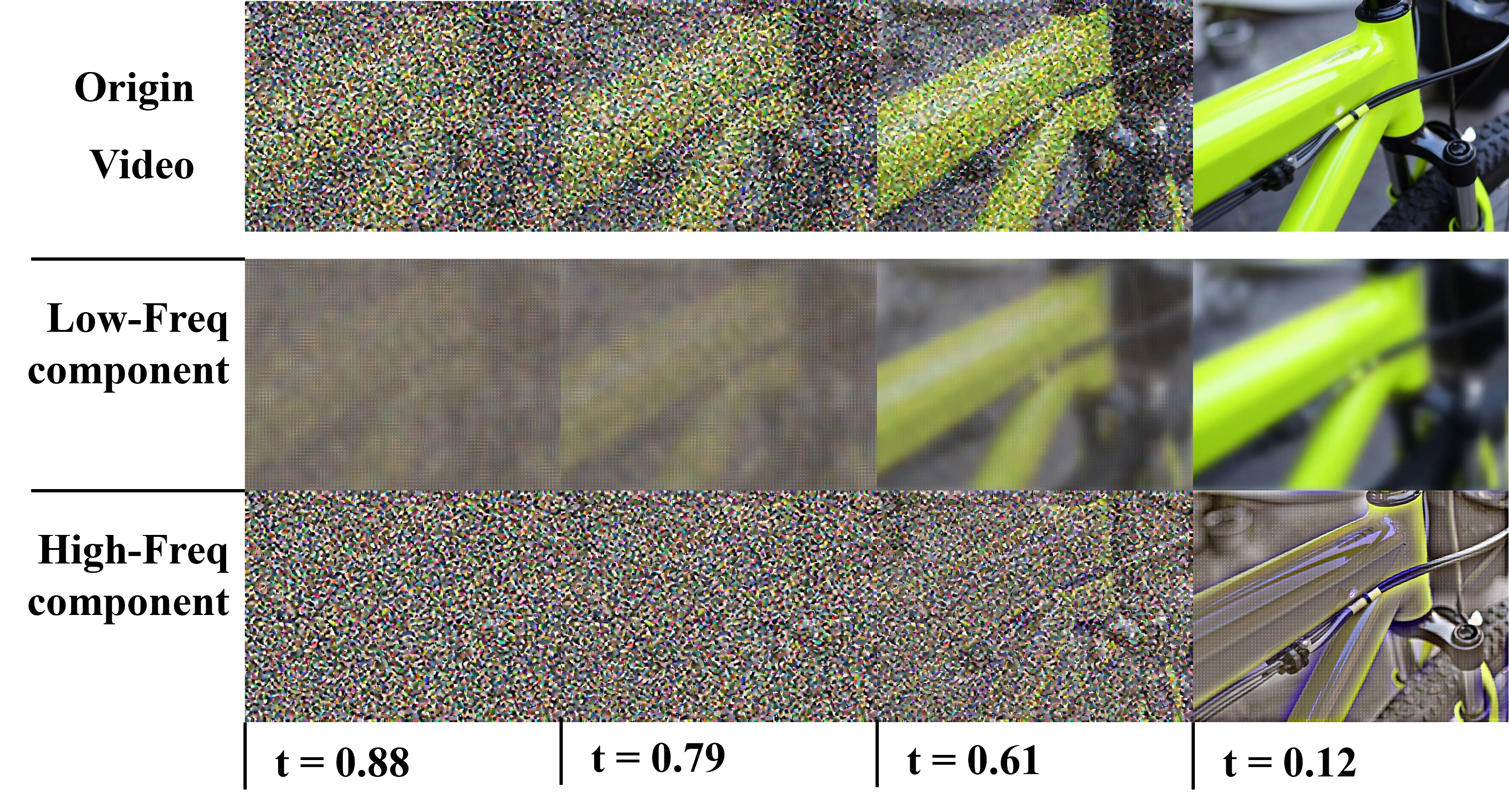}
    \caption{Timestep Hierarchy in Frequency Information Reconstruction. \normalfont{Initially, high-frequency components appear as unstructured noise, but as $t$ decreases, they reveal meaningful structures such as edges and motion details.}}
    \label{fig:high_freq_loss}
\end{figure}

\subsection{High-Frequency Preservation} \label{sec:highfreq}
High-frequency details are crucial for video visual quality, detail richness, and dynamic degree. We introduce a frequency-domain loss that explicitly preserves details. 

For a video latent $z \in \mathbb{R}^{C \times T \times H \times W}$, where $C$ is the channel dimension, $T$ is the temporal dimension, and $H \times W$ are spatial dimensions, we apply a three-dimensional Fast Fourier Transform (FFT), $\mathcal{F}(z)$, to transform it into the frequency domain.
To extract high-frequency components, we design a 3D high-pass filter:
\begin{equation}
H(f_t, f_h, f_w) = 1 - \exp\left(-\alpha \left(\frac{f_t^2}{\sigma_t^2} + \frac{f_h^2}{\sigma_s^2} + \frac{f_w^2}{\sigma_s^2}\right)\right),
\end{equation}
where $\sigma_t$ and $\sigma_s$ are cutoff frequencies for temporal and spatial dimensions respectively, and $\alpha > 0$ controls the steepness of the filter transition. 

The high-frequency components are obtained by element-wise multiplication in the frequency domain:
\begin{equation}
\mathcal{F}_{\text{high}}(z) = \mathcal{F}^{-1}\left(\mathcal{F}(z) \odot H(f_t, f_h, f_w)\right),
\end{equation}
where $\mathcal{F}^{-1}$ denotes the inverse FFT and $\odot$ represents element-wise multiplication.

The high-frequency loss measures the difference between student and teacher outputs in the high-frequency domain:
\begin{equation}
\mathcal{L}_{\text{hf}} = \left\| \mathcal{F}_{\text{high}}(z_\theta) - \mathcal{F}_{\text{high}}(z^*) \right\|^2,
\end{equation}
where $z_\theta$ is the student's output and $z^*$ is the teacher-refined target.

We observe that the generation process exhibits a time-dependent behavior in frequency information reconstruction. As shown in Figure \ref{fig:high_freq_loss}, the high-frequency components initially resemble unstructured noise during the early stages of the diffusion process. As the diffusion timestep $t$ decreases, the video gradually becomes clearer, and the high-frequency components begin to reveal meaningful structures such as edges and motion details. Based on this observation, we introduce a time-aware high-frequency loss that varies with the diffusion timestep $t$ to better align with this hierarchy.
\begin{equation}
\label{eq:high_res_weight}
\lambda(t) = \begin{cases}
\lambda_0, & \text{if stage } = K \text{ and } t \leq 0.5T, \\
0, & \text{otherwise},
\end{cases}
\end{equation}
where $K$ is the final training stage, $T$ is the total diffusion time, and $\lambda_0$ is the base weight. 

The final objective combines trajectory learning with detail preservation:
\begin{equation}
\mathcal{L}^{(k)} = \mathcal{L}_v^{(k)} + \lambda(t) \mathcal{L}_{\text{hf}}.
\end{equation}


This joint optimization ensures robust trajectories through online refinement while maintaining high-frequency details.
Algorithm~\ref{alg:gpd} shows the complete Guided Progressive Distillation procedure. 

\begin{algorithm}[t]
\caption{Guided Progressive Distillation (GPD)}
\label{alg:gpd}
\begin{algorithmic}[1]
\REQUIRE Teacher model $v_\phi$, Total steps $N$, Time steps $\{t_i\}_{i=0}^N$, Student step size $K$, Text dataset $\mathcal{C}$
\ENSURE Fast student model $v_{\theta_K}$
\STATE \textbf{Initialize:} $v_{\theta_1} \leftarrow v_\phi$
\FOR{stage $k = 2$ to $K$}
    \STATE Initialize student $v_{\theta_k}$ from $v_{\theta_{k-1}}$
    \WHILE{not converged}
        \STATE Sample noise $z_{t_N} \sim \mathcal{N}(0, I)$ and prompt $c$ from $\mathcal{C}$
        \FOR{$i = N $ down to $k$ step $-k$}
            \STATE \COMMENT{\textit{1. Online Refinement}}
            \STATE $z_{t_{i-k+1}} \leftarrow \mathcal{S}_{\theta_{k-1}}(z_{t_i}, t_i, t_{i-k+1}, c)$
            \STATE $z_{t_{i-k}}^* \leftarrow \mathcal{S}_\phi(z_{t_{i-k+1}}, t_{i-k+1}, t_{i-k}, c)$
            
            \STATE \COMMENT{\textit{2. Compute Velocity Loss}}
            \STATE $v_{\text{target}} \leftarrow (z_{t_{i-k}}^* - z_{t_i}) / (t_{i-k} - t_i)$
            \STATE $\mathcal{L}_{v} \leftarrow \|v_{\theta_k}(z_{t_i}, t_i, c) - v_{\text{target}}\|^2$
            
            \STATE \COMMENT{\textit{3. Conditional High-Frequency Loss}}
            \STATE $\mathcal{L}_{\text{total}} \leftarrow \mathcal{L}_{v}$
            \IF{$k = K$ \text{and} $i \leq 0.5N$} 
                \STATE $z_{\text{pred}} \leftarrow \mathcal{S}_{\theta_{k}}(z_{t_i}, t_i, t_{i-k}, c)$
                \STATE $\mathcal{L}_{\text{hf}} \leftarrow \|\mathcal{F}_{\text{high}}(z_{\text{pred}}) - \mathcal{F}_{\text{high}}(z_{t_{i-k}}^*)\|^2$
                \STATE $\mathcal{L}_{\text{total}} \leftarrow \mathcal{L}_{\text{total}} + \lambda \cdot \mathcal{L}_{\text{hf}}$
            \ENDIF
            
            \STATE Update $\theta_k \leftarrow \theta_k - \eta \nabla_{\theta_k} \mathcal{L}_{\text{total}}$
            \STATE $z_{t_{i-k}} \leftarrow z_{t_{i-k}}^*$
        \ENDFOR
    \ENDWHILE
\ENDFOR
\RETURN $v_{\theta_K}$
\end{algorithmic}
\end{algorithm}

\begin{table*}[t]
    \centering
    \begin{tabular}{lcc >{\columncolor[gray]{0.90}}c cccccc}
        \toprule
        \parbox{1.5cm}{\textbf{Model}} &
        \parbox{0.5cm}{\centering\small\textbf{Steps}} &
        \parbox{1.1cm}{\centering\small\textbf{H$\times$W$\times$L}} &
        \parbox{1.1cm}{\centering\small\textbf{Total Score}} &
        \parbox{1.1cm}{\centering\small\textbf{Quality Score}} &
        \parbox{1.1cm}{\centering\small\textbf{Semantic Score}} &
        \parbox{1.5cm}{\centering\small\textbf{Background Consistency}} &
        \parbox{1.1cm}{\centering\small\textbf{Aesthetic Quality}} &
        \parbox{1.1cm}{\centering\small\textbf{Multiple Objects}} &
        \parbox{1.1cm}{\centering\small\textbf{Spatial Relationship}} \\
        \midrule

        \multicolumn{10}{l}{\textit{Video Diffusion Models}} \\
        \midrule

        CogVideoX1.5-5B & 48 & 480$\times$720$\times$49   & 82.17 & 82.78 & 79.76 & 97.35 & 62.79 & 69.65 & 80.25 \\
        Open-Sora-2.0   & 48 & 576$\times$1024$\times$120 & 81.71 & 82.10 & 80.14 & 98.75 & 20.74 & 94.50 & 85.98 \\
        Hunyuan Video   & 48 & 720$\times$1280$\times$129 & 83.43 & 85.07 & 76.88 & 97.60 & 60.28 & 66.71 & 72.13 \\
        Step-Video-T2V  & 48 & 922$\times$544$\times$200  & 81.83 & 84.46 & 71.28 & 97.67 & 61.23 & 50.55 & 71.47 \\
        Wan2.1-T2V-1.3B & 48 & 480$\times$832$\times$81   & 83.92 & 85.91 & 75.98 & 97.94 & 69.83 & 75.99 & 74.40 \\
        
        \midrule
        \multicolumn{10}{l}{\textit{Acceleration Pipelines}} \\
        \midrule

        T2V-Turbo-V2    & 16 & 320$\times$512$\times$16   & 83.52 & 85.13 & 77.12 & 96.71 & 62.61 & 61.49 & 43.32 \\
        AccVideo        & 5  & 544$\times$960$\times$72   & 83.28 & 84.58 & 78.06 & 97.43 & 62.08 & 67.39 & 75.73 \\
        CausVid         & 3  & 480$\times$832$\times$81   & 83.65 & 85.97 & 74.38 & 95.07 & 66.28 & 72.03 & 72.08 \\
        PeRFlow         & 6  & 480$\times$832$\times$81   & 82.54 & 84.29 & 75.56 & 98.15 & 70.00 & 79.04 & 75.81 \\
        \textbf{GPD (ours)} & 6 & 480$\times$832$\times$81 & \textbf{84.04} & 85.55 & 77.99 & 98.18 & 71.27 & 78.81 & 82.85 \\
        \textbf{GPD (ours)} & 6  & 720$\times$1280$\times$81  & 82.64 & 84.12 & 76.70 & 97.84 & 69.58 & 74.54 & 84.79 \\
        
        \bottomrule
    \end{tabular}
    \caption{Text-to-video generation comparison.}
    \label{tab:main}
\end{table*}

\section{Experiment}
\subsection{Experiment Setup}
We build GPD acceleration pipeline upon Wan2.1-1.3B diffusion architecture~\cite{wanWanOpenAdvanced2025a}. Both teacher and student models adopt Wan2.1-1.3B architectures and are initialized with the officially released checkpoint. 

For the training configuration, we utilize the textual descriptions from the OpenSora dataset~\cite{zhengOpenSoraDemocratizingEfficient2024,pengOpenSora20Training2025}. The generated video resolution is set to $480 \times 832$ with 81 frames. Training is conducted on 4 NVIDIA A100 (80GB) GPUs with a batch size of 4. We employ the AdamW optimizer with a learning rate of $1e^{-6}$ and a weight decay of $1e^{-2}$. During the progressive distillation process, each stage is trained for 150 iterations. 
For velocity alignment training, following previous works, we apply classifier-free guidance (CFG)~\cite{hoClassifierFreeDiffusionGuidance2022a,luoLatentConsistencyModels2023a} during the prediction of both the teacher and the frozen student models. The CFG scale is linearly decayed from 6.0 to 1.5 as the training stages progress.
Regarding the high-frequency loss, the cutoff frequencies are set to $\sigma_t = \sigma_s = 0.25$, and $\lambda_0 = 0.5$.

Besides, we trained the Wan2.1-1.3B model using the PeRFlow and CausVid methods for comparison. These experiments were conducted on the same 4 NVIDIA A100 (80GB) GPUs, adhering to the hyperparameter settings declared in their original papers. Regarding the training data, CausVid utilizes real-world web videos. In contrast, PeRFlow employs synthetic data generated by the teacher model, for which we utilized the same prompt dataset as our method.

\begin{table}[t]
\centering
\begin{tabular}{l c c c c}
\toprule
 & \textbf{Video} & \multicolumn{3}{c}{\textbf{Cost (GPU-days)}} \\
\cmidrule(lr){3-5}
\textbf{Method} & \textbf{Dataset} & \textbf{Synthesis} & \textbf{Train} & \textbf{Total} \\
\midrule
PeRFlow              & yes & 0.729 & 0.190 & 0.919 \\[2pt]
CausVid              & yes & --    & 1.354 & 1.354 \\[2pt]
\textbf{GPD (ours)}  & \textbf{no}  & --    & 0.550 & \textbf{0.550} \\[2pt]
\bottomrule
\end{tabular}
\caption{Comparison of training cost required by different methods.}
\label{tab:gpu_days_comparison}
\end{table}

\subsection{Comparison on training efficiency}
Table~\ref{tab:gpu_days_comparison} presents a quantitative comparison of the training costs across different methods. As shown, our GPD method achieves the lowest computational cost, requiring significantly fewer GPU-days than both PeRFlow and CausVid. 

This superior efficiency is attributed to two key factors. First, our training approach is data-efficient; it relies solely on textual prompts, thereby eliminating the overhead associated with pre-synthesizing video data (as required by PeRFlow) or preprocessing large-scale web videos (as required by CausVid). Second, the proposed progressive distillation strategy ensures rapid convergence. As mentioned in the setup, each stage requires only 150 iterations, which substantially reduces the overall training duration.

\begin{figure*}[!t]
    \centering
    \includegraphics[width=\textwidth]{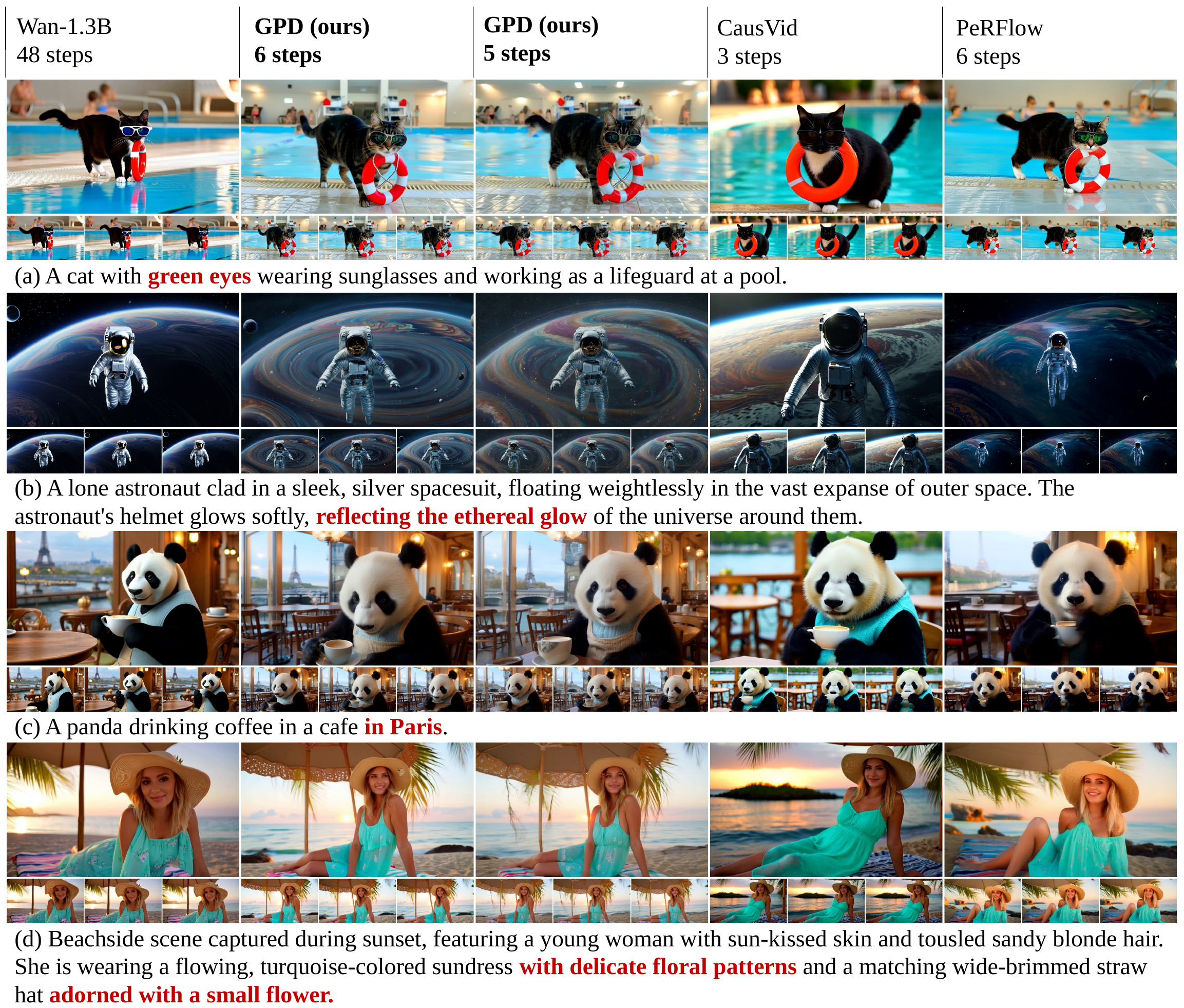}
    \caption{Qualitative results on text-to-video. \normalfont{Qualitative comparison showing our method achieves 8× faster inference than Wan2.1-1.3B while maintaining superior visual quality. Compared to PeRFlow, our approach eliminates noticeable distortions in subjects and backgrounds. Compared to CausVid, our method maintains stronger semantic coherence and preserves fine-grained elements. }}
    \label{fig:post}
\end{figure*}

\subsection{Evaluation on Text-to-Video Generation}
We evaluate our method on VBench~\cite{huangVBenchComprehensiveBenchmark2023}, a comprehensive video generation benchmark comprising 16 metrics designed to systematically assess both video quality and semantic alignment.

We compare our GPD model with both standard text-to-video generation baselines and fast-generation approaches. Among traditional full-step generative models, we evaluate against CogVideoX~\cite{hongCogVideoLargescalePretraining2022,yangCogVideoXTexttoVideoDiffusion2025}, OpenSora~\cite{zhengOpenSoraDemocratizingEfficient2024,pengOpenSora20Training2025}, and HunyuanVideo~\cite{kongHunyuanVideoSystematicFramework2025}, all of which represent state-of-the-art transformer-based methods. For acceleration baselines, we include T2V-Turbo-V2~\cite{liT2VTurbov2EnhancingVideo2024},  AccVideo~\cite{zhangAccVideoAcceleratingVideo2025a}, PeRFlow~\cite{yanPeRFlowPiecewiseRectified2024b} and CausVid~\cite{yinSlowBidirectionalFast2025}. In particular, PeRFlow and CausVid employ the same base model, training video resolution, and frame count as our method.
For our GPD model, we generate videos at two resolutions, 480p and 720p, with a fixed frame length of 81 in both cases.

Compared to the base Wan2.1-1.3B model, we achieve competitive performance while improving inference efficiency by 8$\times$, significantly enhancing practical usability. 
Our method completely outperforms the PeRFlow model, as is clearly shown in Table \ref{tab:main}. 
While CausVid operates with fewer inference steps, it suffers from degraded text-alignment and loses fine-grained elements in the generated videos. In contrast, our method maintains strong semantic coherence, as shown in Figure \ref{fig:post}. 

Our method demonstrates strong generalization across different resolutions; models distilled at lower resolutions still maintain high quality when generating high-resolution videos. The slight decline in the total score for 720p video generation compared to 480p is attributed to the base model, Wan2.1-T2V-1.3B, lacking strong native support for 720p videos.


\begin{figure*}[!t]
\centering
    \includegraphics[width=\linewidth]{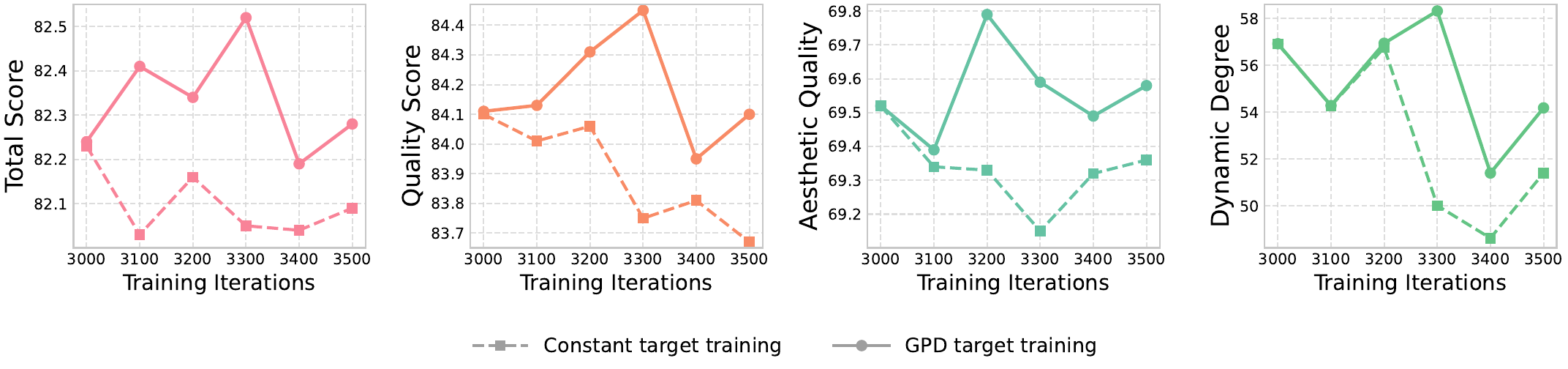}
    
    \begingroup
    \setlength{\abovecaptionskip}{2pt}
    \setlength{\belowcaptionskip}{0pt}
    \caption{Online refinement training improves distillation quality. Compared with conventional training using a constant target, online refinement post-training yields consistent gains in the overall score and across multiple sub-metrics, indicating improved visual fidelity and stronger temporal consistency.}\label{fig:ablation_refine}
    \endgroup

\end{figure*}

\section{Ablation Studies}

\subsection{Effect of Online Teacher Guidance}
We conduct an ablation study to evaluate the effectiveness of teacher guidance. We first train the student model with constant targets derived from the teacher trajectory; after 3,000 training steps, the model converges and reaches its highest total score. Starting from this checkpoint, we branch into two runs: (i) continuing with constant-target training, and (ii) switching to teacher-refined (GPD) targets. All other training settings are kept identical, and only the training targets differ. Figure~\ref{fig:ablation_refine} reports the evolution of VBench total score together with several representative sub-metrics from 3,000 to 3,500 iterations.

Overall, the GPD-target branch consistently outperforms the constant-target branch, reaching its highest peak total score around 3,300 iterations. It also improves quality and aesthetics and, most notably, better preserves motion dynamics than constant targets. These results suggest that teacher-refined targets provide more informative and corrective supervision than static constant targets.


\begin{table}[t]
    \centering
        \begin{tabular}{lcccc}
        \toprule
        \parbox{1.2cm}{\textbf{Model}} & 
        \parbox{1.2cm}{\centering\small\textbf{Inference Steps}} & 
        \parbox{1.0cm}{\centering\small\textbf{Total Score}} & 
        \parbox{1.0cm}{\centering\small\textbf{Semantic Score}} & 
        \parbox{1.0cm}{\centering\small\textbf{Quality Score}} \\
        \midrule
        Wan2.1-T2V & 48 & 83.92 & 75.98 & 85.91 \\
        \midrule
        \multirow{6}{*}{GPD} & 24 & \underline{83.94} & 76.64 & 85.77 \\
        & 16 & \underline{83.69} & 76.72 & 85.43 \\
        & 12 & \underline{83.84} & 76.34 & 85.71 \\
        & 8 & \underline{84.13} & 77.23 & 85.85 \\
        & 6 & \underline{84.04} & 77.99 & 85.55 \\
        & 5 & 83.60 & 77.22 & 85.20 \\
        
        \bottomrule
        \end{tabular}
    \caption{\normalfont{Performance across different inference steps.}}
    \label{tab:inference_steps_comparison}
\end{table}

\subsection{Progressive Training Analysis}

Progressive training is essential for managing the complexity of trajectory learning as step size increases. By gradually increasing the step size across multiple stages, our progressive approach allows the model to incrementally learn longer predictions while maintaining stable performance at each stage.

Table~\ref{tab:inference_steps_comparison} demonstrates the effectiveness of our progressive training strategy across different inference step counts. The results show that our method maintains or even improves performance as the number of inference steps decreases from 24 to 6. Notably, the semantic score consistently improves from 75.98 to 77.99, while the quality score remains stable around 85.5-85.9. The optimal performance is achieved at 6-8 steps, where the model strikes the best balance between generation speed and output quality.

\begin{table}
  \centering
  \begin{tabular}{lcc}
      \toprule
      & 
      \parbox{1.5cm}{\centering\small\textbf{Dynamic degree}} & 
      \parbox{1.5cm}{\centering\small\textbf{Image quality}} \\
      
      \midrule
      No high-freq loss & 54.17 & 68.96 \\
      Const high-freq loss & 55.56 & 69.04 \\
      Weighted high-freq loss & \textbf{56.94} & \textbf{69.07} \\
      
      \bottomrule
  \end{tabular}
  \caption{\normalfont{Ablation on High Frequency Loss.}}
  \label{tab:high_freq_loss}
\end{table}

\subsection{High-Frequency Loss Analysis}

High-frequency components are critical for capturing motion dynamics and preserving visual quality in videos.
We compared three high-frequency loss configurations for training GPD: no high-frequency loss, constant high-frequency loss, and weighted high-frequency loss. The model trained without high-frequency loss yielded the lowest dynamic degree and image quality, primarily due to a lack of alignment with high-frequency content. The constant high-frequency loss, which employs a fixed weight across diffusion timesteps, achieved improved scores.

Our weighted high-frequency loss applies a timestep-adapted high-frequency loss, dynamically adjusting the loss weight according to the diffusion timestep, as described in Section~\ref{sec:highfreq}. As shown in Table~\ref{tab:high_freq_loss}, this strategy achieves the best performance on both dynamic degree and image quality. We attribute the improvement to the fact that high-frequency reconstruction is highly sensitive to diffusion timesteps; the proposed weighting scheme better matches the stage-dependent nature of the high-frequency formation process.

\section{Conclusion}
We introduced Guided Progressive Distillation (GPD), a novel framework for accelerating video diffusion models while maintaining high-quality outputs. By combining online refinement, progressive distillation, and high-frequency preservation, GPD achieves an 8$\times$ speedup, reducing video generation from 48 steps to just 6 steps, while outperforming existing methods on the VBench benchmark with a total score of 84.04\%. Our approach eliminates the need for pre-collected datasets, ensures robust trajectory learning, and preserves fine-grained details, offering a practical solution for efficient, high-quality video synthesis.

\bibliographystyle{named}
\bibliography{ijcai26}

\begin{thebibliography}{}

\bibitem[\protect\citeauthoryear{Blattmann \bgroup \em et al.\egroup }{2023}]{blattmannStableVideoDiffusion2023}
Andreas Blattmann, Tim Dockhorn, Sumith Kulal, Daniel Mendelevitch, Maciej Kilian, Dominik Lorenz, Yam Levi, Zion English, Vikram Voleti, Adam Letts, Varun Jampani, and Robin Rombach.
\newblock Stable video diffusion: Scaling latent video diffusion models to large datasets, November 2023.

\bibitem[\protect\citeauthoryear{Chadebec \bgroup \em et al.\egroup }{2024}]{chadebecFlashDiffusionAccelerating2024}
Cl{\'e}ment Chadebec, Onur Tasar, Eyal Benaroche, and Benjamin Aubin.
\newblock Flash diffusion: Accelerating any conditional diffusion model for few steps image generation, December 2024.

\bibitem[\protect\citeauthoryear{Esser \bgroup \em et al.\egroup }{2024}]{esserScalingRectifiedFlow2024a}
Patrick Esser, Sumith Kulal, Andreas Blattmann, Rahim Entezari, Jonas M{\"u}ller, Harry Saini, Yam Levi, Dominik Lorenz, Axel Sauer, Frederic Boesel, Dustin Podell, Tim Dockhorn, Zion English, Kyle Lacey, Alex Goodwin, Yannik Marek, and Robin Rombach.
\newblock Scaling rectified flow transformers for high-resolution image synthesis, March 2024.

\bibitem[\protect\citeauthoryear{Ho and Salimans}{2022}]{hoClassifierFreeDiffusionGuidance2022a}
Jonathan Ho and Tim Salimans.
\newblock Classifier-free diffusion guidance, July 2022.

\bibitem[\protect\citeauthoryear{Ho \bgroup \em et al.\egroup }{2020}]{hoDenoisingDiffusionProbabilistic2020b}
Jonathan Ho, Ajay Jain, and Pieter Abbeel.
\newblock Denoising diffusion probabilistic models, December 2020.

\bibitem[\protect\citeauthoryear{Hong \bgroup \em et al.\egroup }{2022}]{hongCogVideoLargescalePretraining2022}
Wenyi Hong, Ming Ding, Wendi Zheng, Xinghan Liu, and Jie Tang.
\newblock Cogvideo: Large-scale pretraining for text-to-video generation via transformers, May 2022.

\bibitem[\protect\citeauthoryear{Huang \bgroup \em et al.\egroup }{2023}]{huangVBenchComprehensiveBenchmark2023}
Ziqi Huang, Yinan He, Jiashuo Yu, Fan Zhang, Chenyang Si, Yuming Jiang, Yuanhan Zhang, Tianxing Wu, Qingyang Jin, Nattapol Chanpaisit, Yaohui Wang, Xinyuan Chen, Limin Wang, Dahua Lin, Yu~Qiao, and Ziwei Liu.
\newblock Vbench: Comprehensive benchmark suite for video generative models, November 2023.

\bibitem[\protect\citeauthoryear{Ke \bgroup \em et al.\egroup }{2025}]{keProReflowProgressiveReflow2025}
Lei Ke, Haohang Xu, Xuefei Ning, Yu~Li, Jiajun Li, Haoling Li, Yuxuan Lin, Dongsheng Jiang, Yujiu Yang, and Linfeng Zhang.
\newblock Proreflow: Progressive reflow with decomposed velocity, March 2025.

\bibitem[\protect\citeauthoryear{Kong \bgroup \em et al.\egroup }{2025}]{kongHunyuanVideoSystematicFramework2025}
Weijie Kong, Qi~Tian, Zijian Zhang, Rox Min, Zuozhuo Dai, Jin Zhou, Jiangfeng Xiong, Xin Li, Bo~Wu, Jianwei Zhang, Kathrina Wu, Qin Lin, Junkun Yuan, Yanxin Long, Aladdin Wang, Andong Wang, Changlin Li, Duojun Huang, Fang Yang, Hao Tan, Hongmei Wang, Jacob Song, Jiawang Bai, Jianbing Wu, Jinbao Xue, Joey Wang, Kai Wang, Mengyang Liu, Pengyu Li, Shuai Li, Weiyan Wang, Wenqing Yu, Xinchi Deng, Yang Li, Yi~Chen, Yutao Cui, Yuanbo Peng, Zhentao Yu, Zhiyu He, Zhiyong Xu, Zixiang Zhou, Zunnan Xu, Yangyu Tao, Qinglin Lu, Songtao Liu, Dax Zhou, Hongfa Wang, Yong Yang, Di~Wang, Yuhong Liu, Jie Jiang, and Caesar Zhong.
\newblock Hunyuanvideo: A systematic framework for large video generative models, January 2025.

\bibitem[\protect\citeauthoryear{Li \bgroup \em et al.\egroup }{2024}]{liT2VTurbov2EnhancingVideo2024}
Jiachen Li, Qian Long, Jian Zheng, Xiaofeng Gao, Robinson Piramuthu, Wenhu Chen, and William~Yang Wang.
\newblock T2v-turbo-v2: Enhancing video generation model post-training through data, reward, and conditional guidance design, October 2024.

\bibitem[\protect\citeauthoryear{Lin \bgroup \em et al.\egroup }{2024}]{linSDXLLightningProgressiveAdversarial2024}
Shanchuan Lin, Anran Wang, and Xiao Yang.
\newblock Sdxl-lightning: Progressive adversarial diffusion distillation, March 2024.

\bibitem[\protect\citeauthoryear{Lin \bgroup \em et al.\egroup }{2025}]{linDiffusionAdversarialPostTraining2025}
Shanchuan Lin, Xin Xia, Yuxi Ren, Ceyuan Yang, Xuefeng Xiao, and Lu~Jiang.
\newblock Diffusion adversarial post-training for one-step video generation, January 2025.

\bibitem[\protect\citeauthoryear{Lipman \bgroup \em et al.\egroup }{2023}]{lipmanFlowMatchingGenerative2023a}
Yaron Lipman, Ricky T.~Q. Chen, Heli {Ben-Hamu}, Maximilian Nickel, and Matt Le.
\newblock Flow matching for generative modeling, February 2023.

\bibitem[\protect\citeauthoryear{Liu \bgroup \em et al.\egroup }{2022}]{liuFlowStraightFast2022a}
Xingchao Liu, Chengyue Gong, and Qiang Liu.
\newblock Flow straight and fast: Learning to generate and transfer data with rectified flow, September 2022.

\bibitem[\protect\citeauthoryear{Liu \bgroup \em et al.\egroup }{2024}]{liuInstaFlowOneStep2024}
Xingchao Liu, Xiwen Zhang, Jianzhu Ma, Jian Peng, and Qiang Liu.
\newblock Instaflow: One step is enough for high-quality diffusion-based text-to-image generation, March 2024.

\bibitem[\protect\citeauthoryear{Luo \bgroup \em et al.\egroup }{2023a}]{luoLatentConsistencyModels2023a}
Simian Luo, Yiqin Tan, Longbo Huang, Jian Li, and Hang Zhao.
\newblock Latent consistency models: Synthesizing high-resolution images with few-step inference, October 2023.

\bibitem[\protect\citeauthoryear{Luo \bgroup \em et al.\egroup }{2023b}]{luoLCMLoRAUniversalStableDiffusion2023}
Simian Luo, Yiqin Tan, Suraj Patil, Daniel Gu, Patrick von Platen, Apolin{\'a}rio Passos, Longbo Huang, Jian Li, and Hang Zhao.
\newblock Lcm-lora: A universal stable-diffusion acceleration module, November 2023.

\bibitem[\protect\citeauthoryear{Luo \bgroup \em et al.\egroup }{2024}]{luoOneStepDiffusionDistillation2024}
Weijian Luo, Zemin Huang, Zhengyang Geng, J.~Zico Kolter, and Guo-jun Qi.
\newblock One-step diffusion distillation through score implicit matching, October 2024.

\bibitem[\protect\citeauthoryear{Ma \bgroup \em et al.\egroup }{2025}]{maStepVideoT2VTechnicalReport2025}
Guoqing Ma, Haoyang Huang, Kun Yan, Liangyu Chen, Nan Duan, Shengming Yin, Changyi Wan, Ranchen Ming, Xiaoniu Song, Xing Chen, Yu~Zhou, Deshan Sun, Deyu Zhou, Jian Zhou, Kaijun Tan, Kang An, Mei Chen, Wei Ji, Qiling Wu, Wen Sun, Xin Han, Yanan Wei, Zheng Ge, Aojie Li, Bin Wang, Bizhu Huang, Bo~Wang, Brian Li, Changxing Miao, Chen Xu, Chenfei Wu, Chenguang Yu, Dapeng Shi, Dingyuan Hu, Enle Liu, Gang Yu, Ge~Yang, Guanzhe Huang, Gulin Yan, Haiyang Feng, Hao Nie, Haonan Jia, Hanpeng Hu, Hanqi Chen, Haolong Yan, Heng Wang, Hongcheng Guo, Huilin Xiong, Huixin Xiong, Jiahao Gong, Jianchang Wu, Jiaoren Wu, Jie Wu, Jie Yang, Jiashuai Liu, Jiashuo Li, Jingyang Zhang, Junjing Guo, Junzhe Lin, Kaixiang Li, Lei Liu, Lei Xia, Liang Zhao, Liguo Tan, Liwen Huang, Liying Shi, Ming Li, Mingliang Li, Muhua Cheng, Na~Wang, Qiaohui Chen, Qinglin He, Qiuyan Liang, Quan Sun, Ran Sun, Rui Wang, Shaoliang Pang, Shiliang Yang, Sitong Liu, Siqi Liu, Shuli Gao, Tiancheng Cao, Tianyu Wang, Weipeng Ming, Wenqing He, Xu~Zhao, Xuelin Zhang,
  Xianfang Zeng, Xiaojia Liu, Xuan Yang, Yaqi Dai, Yanbo Yu, Yang Li, Yineng Deng, Yingming Wang, Yilei Wang, Yuanwei Lu, Yu~Chen, Yu~Luo, Yuchu Luo, Yuhe Yin, Yuheng Feng, Yuxiang Yang, Zecheng Tang, Zekai Zhang, Zidong Yang, Binxing Jiao, Jiansheng Chen, Jing Li, Shuchang Zhou, Xiangyu Zhang, Xinhao Zhang, Yibo Zhu, Heung-Yeung Shum, and Daxin Jiang.
\newblock Step-video-t2v technical report: The practice, challenges, and future of video foundation model, February 2025.

\bibitem[\protect\citeauthoryear{Peng \bgroup \em et al.\egroup }{2025}]{pengOpenSora20Training2025}
Xiangyu Peng, Zangwei Zheng, Chenhui Shen, Tom Young, Xinying Guo, Binluo Wang, Hang Xu, Hongxin Liu, Mingyan Jiang, Wenjun Li, Yuhui Wang, Anbang Ye, Gang Ren, Qianran Ma, Wanying Liang, Xiang Lian, Xiwen Wu, Yuting Zhong, Zhuangyan Li, Chaoyu Gong, Guojun Lei, Leijun Cheng, Limin Zhang, Minghao Li, Ruijie Zhang, Silan Hu, Shijie Huang, Xiaokang Wang, Yuanheng Zhao, Yuqi Wang, Ziang Wei, and Yang You.
\newblock Open-sora 2.0: Training a commercial-level video generation model in \$200k, March 2025.

\bibitem[\protect\citeauthoryear{Salimans and Ho}{2022}]{salimansProgressiveDistillationFast2022a}
Tim Salimans and Jonathan Ho.
\newblock Progressive distillation for fast sampling of diffusion models, June 2022.

\bibitem[\protect\citeauthoryear{Sauer \bgroup \em et al.\egroup }{2023}]{sauerAdversarialDiffusionDistillation2023}
Axel Sauer, Dominik Lorenz, Andreas Blattmann, and Robin Rombach.
\newblock Adversarial diffusion distillation, November 2023.

\bibitem[\protect\citeauthoryear{Song \bgroup \em et al.\egroup }{2021}]{songScoreBasedGenerativeModeling2021}
Yang Song, Jascha {Sohl-Dickstein}, Diederik~P. Kingma, Abhishek Kumar, Stefano Ermon, and Ben Poole.
\newblock Score-based generative modeling through stochastic differential equations, February 2021.

\bibitem[\protect\citeauthoryear{Song \bgroup \em et al.\egroup }{2023}]{songConsistencyModels2023a}
Yang Song, Prafulla Dhariwal, Mark Chen, and Ilya Sutskever.
\newblock Consistency models, May 2023.

\bibitem[\protect\citeauthoryear{Wan \bgroup \em et al.\egroup }{2025}]{wanWanOpenAdvanced2025a}
Team Wan, Ang Wang, Baole Ai, Bin Wen, Chaojie Mao, Chen-Wei Xie, Di~Chen, Feiwu Yu, Haiming Zhao, Jianxiao Yang, Jianyuan Zeng, Jiayu Wang, Jingfeng Zhang, Jingren Zhou, Jinkai Wang, Jixuan Chen, Kai Zhu, Kang Zhao, Keyu Yan, Lianghua Huang, Mengyang Feng, Ningyi Zhang, Pandeng Li, Pingyu Wu, Ruihang Chu, Ruili Feng, Shiwei Zhang, Siyang Sun, Tao Fang, Tianxing Wang, Tianyi Gui, Tingyu Weng, Tong Shen, Wei Lin, Wei Wang, Wei Wang, Wenmeng Zhou, Wente Wang, Wenting Shen, Wenyuan Yu, Xianzhong Shi, Xiaoming Huang, Xin Xu, Yan Kou, Yangyu Lv, Yifei Li, Yijing Liu, Yiming Wang, Yingya Zhang, Yitong Huang, Yong Li, You Wu, Yu~Liu, Yulin Pan, Yun Zheng, Yuntao Hong, Yupeng Shi, Yutong Feng, Zeyinzi Jiang, Zhen Han, Zhi-Fan Wu, and Ziyu Liu.
\newblock Wan: Open and advanced large-scale video generative models, April 2025.

\bibitem[\protect\citeauthoryear{Wang \bgroup \em et al.\egroup }{2023}]{wangVideoLCMVideoLatent2023}
Xiang Wang, Shiwei Zhang, Han Zhang, Yu~Liu, Yingya Zhang, Changxin Gao, and Nong Sang.
\newblock Videolcm: Video latent consistency model, December 2023.

\bibitem[\protect\citeauthoryear{Wang \bgroup \em et al.\egroup }{2024}]{wangMagicVideoV2MultiStageHighAesthetic2024}
Weimin Wang, Jiawei Liu, Zhijie Lin, Jiangqiao Yan, Shuo Chen, Chetwin Low, Tuyen Hoang, Jie Wu, Jun~Hao Liew, Hanshu Yan, Daquan Zhou, and Jiashi Feng.
\newblock Magicvideo-v2: Multi-stage high-aesthetic video generation, January 2024.

\bibitem[\protect\citeauthoryear{Yan \bgroup \em et al.\egroup }{2024}]{yanPeRFlowPiecewiseRectified2024b}
Hanshu Yan, Xingchao Liu, Jiachun Pan, Jun~Hao Liew, Qiang Liu, and Jiashi Feng.
\newblock Perflow: Piecewise rectified flow as universal plug-and-play accelerator, September 2024.

\bibitem[\protect\citeauthoryear{Yang \bgroup \em et al.\egroup }{2025}]{yangCogVideoXTexttoVideoDiffusion2025}
Zhuoyi Yang, Jiayan Teng, Wendi Zheng, Ming Ding, Shiyu Huang, Jiazheng Xu, Yuanming Yang, Wenyi Hong, Xiaohan Zhang, Guanyu Feng, Da~Yin, Yuxuan Zhang, Weihan Wang, Yean Cheng, Bin Xu, Xiaotao Gu, Yuxiao Dong, and Jie Tang.
\newblock Cogvideox: Text-to-video diffusion models with an expert transformer, March 2025.

\bibitem[\protect\citeauthoryear{Yin \bgroup \em et al.\egroup }{2024a}]{yinImprovedDistributionMatching2024}
Tianwei Yin, Micha{\"e}l Gharbi, Taesung Park, Richard Zhang, Eli Shechtman, Fredo Durand, and William~T. Freeman.
\newblock Improved distribution matching distillation for fast image synthesis, May 2024.

\bibitem[\protect\citeauthoryear{Yin \bgroup \em et al.\egroup }{2024b}]{yinOnestepDiffusionDistribution2024a}
Tianwei Yin, Micha{\"e}l Gharbi, Richard Zhang, Eli Shechtman, Fredo Durand, William~T. Freeman, and Taesung Park.
\newblock One-step diffusion with distribution matching distillation, October 2024.

\bibitem[\protect\citeauthoryear{Yin \bgroup \em et al.\egroup }{2025}]{yinSlowBidirectionalFast2025}
Tianwei Yin, Qiang Zhang, Richard Zhang, William~T. Freeman, Fredo Durand, Eli Shechtman, and Xun Huang.
\newblock From slow bidirectional to fast autoregressive video diffusion models, January 2025.

\bibitem[\protect\citeauthoryear{Zhang \bgroup \em et al.\egroup }{2025}]{zhangAccVideoAcceleratingVideo2025a}
Haiyu Zhang, Xinyuan Chen, Yaohui Wang, Xihui Liu, Yunhong Wang, and Yu~Qiao.
\newblock Accvideo: Accelerating video diffusion model with synthetic dataset, March 2025.

\bibitem[\protect\citeauthoryear{Zheng \bgroup \em et al.\egroup }{2024}]{zhengOpenSoraDemocratizingEfficient2024}
Zangwei Zheng, Xiangyu Peng, Tianji Yang, Chenhui Shen, Shenggui Li, Hongxin Liu, Yukun Zhou, Tianyi Li, and Yang You.
\newblock Open-sora: Democratizing efficient video production for all, December 2024.

\bibitem[\protect\citeauthoryear{Zhou \bgroup \em et al.\egroup }{2023}]{zhouMagicVideoEfficientVideo2023}
Daquan Zhou, Weimin Wang, Hanshu Yan, Weiwei Lv, Yizhe Zhu, and Jiashi Feng.
\newblock Magicvideo: Efficient video generation with latent diffusion models, May 2023.

\end{thebibliography}

\end{document}